\documentclass[sigconf,natbib=true,anonymous=false]{acmart}
\AtBeginDocument{%
  \providecommand\BibTeX{{%
    \normalfont B\kern-0.5em{\scshape i\kern-0.25em b}\kern-0.8em\TeX}}}

\renewcommand\footnotetextcopyrightpermission[1]{} 
\makeatletter
\renewcommand\@formatdoi[1]{\ignorespaces}
\makeatother

\usepackage{graphicx}
\usepackage{subfig}
\usepackage{adjustbox}
\usepackage{hyperref}
\usepackage{titlesec}
\titlespacing*{\subsection}{0pt}{0.5ex plus 1ex minus 0.2ex}{.5ex plus 0.2ex}

\begin{document}

\title{Enhancing News Summarization with ELearnFit through Efficient In-Context Learning and Efficient Fine-Tuning}

\author{Che Guan}
\email{che.guan@alliancebernstein.com}
\author{Andrew Chin}
\email{andrew.chin@alliancebernstein.com}
\affiliation{%
  \institution{AllianceBernstein}
  \streetaddress{1345 6th Ave}
  \city{New York}
  \state{NY}
  \country{USA}
  \postcode{10105}
}

\author{Puya Vahabi}
\affiliation{%
  \institution{University of California, Berkeley}
  \streetaddress{University Avenue and, Oxford St}
  \city{Berkeley}
  \state{CA}
  \country{USA}}
  \postcode{94720}
\email{puya@ischool.berkeley.edu}


\begin{abstract}
With the deluge of information delivered by the daily news cycle, there is a growing need to effectively and efficiently summarize news feeds for quick consumption. We leverage large language models (LLMs), with their advanced learning and generative abilities as compared to conventional language models, to generate concise and coherent summaries for news articles from the XSum dataset. Our paper focuses on two key aspects of LLMs – Efficient in-context Learning (ELearn) and Parameter Efficient Fine-tuning (EFit). Under ELearn, we find that increasing the number of shots in prompts and utilizing simple templates generally improve the quality of summaries. We also find that utilizing relevant examples in few-shot learning for ELearn does not improve model performance. In addition, we studied EFit using different methods and demonstrate that fine-tuning the first layer of LLMs produces better outcomes as compared to fine-tuning other layers or utilizing LoRA. We also find that leveraging more relevant training samples using selective layers does not result in better performance. By combining ELearn and EFit, we create a new model (ELearnFit) that leverages the benefits of both few-shot learning and fine-tuning, and produces superior performance to either model alone. We also use ELearnFit to highlight the trade-offs between prompting and fine-tuning, especially for situations where only a limited number of annotated samples are available. Ultimately, our research provides practical techniques to optimize news summarization during the prompting and fine-tuning stages, and enhances the synthesis of news articles.
\end{abstract}





\maketitle
\pagestyle{plain}

\section{Introduction}
\label{sec:Introduction}
There has been an overload of information with each passing day – data is more voluminous, comes in more varieties and arrives at higher velocity. The news cycle is a good example of this trend, making it more difficult to read and synthesize the vast amount of information coming our way. The advent of large language models (LLMs) has led to a substantial improvement in the effectiveness and comprehensibility of news summarization. LLMs present two ways to address downstream tasks – through prompt engineering and fine-tuning. In our research, we explore various techniques to improve model performance through better prompts and fine-tuning methods. 

First, we study efficient in-context learning, which we call ELearn to denote the process of the model learning through prompts. We examine the impact of LLM size, the number of shots, and various templates during the in-context learning . We also select relevant samples in prompting in an attempt to improve performance. We then explore efficient methods to fine-tune LLMs. Calling this technique EFit, we test the performance of selective layer fine-tuning and LoRA in news summarization. We also utilize selective samples to improve the training set for the fine-tuning process.  Finally, we combine ELearn and EFit to create ELearnFit and find that this model achieves superior performance versus either model alone. 

We make various contributions to existing research on news summarization \footnote{The codes used in this study were derived from the class "Deep Multi-Task and Meta Learning" offered by Stanford School of Engineering. We implemented and adapted these foundational project codes to meet the specific requirements of our study.}. Through ELearn, we find that using larger models, increasing the number of shots during prompting, and leveraging simple templates can all enhance model performance. We also show that utilizing selective relevant examples during prompting does not meaningfully impact performance. Through EFit, we find that fine-tuning the first layer of LLMs produces better outcomes as compared to fine-tuning other layers or utilizing LoRA, and leveraging more relevant training samples using selective samples does not result in better performance. The combined model, ELearnFit, leverages the best of both worlds and suggests practical implementations for practitioners, especially when using a limited number of annotated samples.  

\section{Related Work}
\label{sec:reference}
The evolution of news summarization techniques has been driven by advancements in NLP and the increasing availability of large-scale datasets. Early news summarization techniques relied on statistical methods, such as frequency analysis and clustering, to extract important information from news articles. These methods were limited in their ability to capture the semantics and context of the news content.  

With the advent of deep learning, news summarization techniques have undergone a significant transformation. Deep learning models, particularly transformer-based architectures such as BERT and GPT-3\cite{vaswani2017attention,devlin2019bert,brown2020language}, have demonstrated remarkable performance in various NLP tasks, including news summarization \cite{goyal2023news}. These models are able to learn complex representations of news articles and generate summaries that are both informative and coherent.  Recent research in news summarization has focused on developing techniques that can handle diverse types of news articles, including long and complex articles, and generate summaries that are tailored to specific user needs and preferences. Additionally, there has been growing interest in explainable news summarization \cite{haonan2020exploring, mahmoudi2023exploring}, which aims to provide users with insights into how summaries are generated and the rationale behind the selection of specific sentences or phrases.

Fine-tuning LLMs and in-context learning are two powerful techniques that have been successfully applied to summarization \cite{fetahu2023instructpts, zhang2023benchmarking,basyal2023text}. Fine-tuning LLMs \cite{tian2023finetuning} involves adapting the pre-trained LLM to the specific task of news summarization by fine-tuning its parameters on a smaller, task-specific dataset. This allows the LLM to leverage its learned knowledge and adapt it to the task of generating informative and coherent summaries. 

In-context learning is a technique where a pre-trained LM utilizes text input to define a task. By providing the model with an instruction and/or a few task demonstrations, it gains the ability to predict subsequent steps and complete additional instances of the task \cite{brown2020language}.  Furthermore, in-context learning can be viewed as a form of implicit Bayesian inference \cite{xie2022explanation}. The model learns to infer a latent concept from the context and uses it to generate a response. The pretraining distribution can be seen as a mixture of hidden Markov models (HMMs), where each HMM represents a different concept. When prompted with a specific context, the model implicitly infers the latent concept that is most relevant to the context and generates a response based on that concept.

To facilitate the training and evaluation of news summarization models, large-scale datasets such as CNN/Daily Mail and XSum \cite{narayan2018dont,cao2022survey} have proven invaluable.  These datasets provide a diverse collection of news articles and human-generated summaries, enabling researchers to benchmark different summarization techniques and track progress in the field.

\section{Approach}
\label{sec:Approach}
In this study, we utilize the XSum dataset, a large-scale collection of news articles with annotated summarizations, as analyzed in Subsection \ref{subsec:Analysis} , to explore various methods for enhancing prompting (ELearn) and fine-tuning (EFit), which will be explained in Subsections \ref{subsec:ELearn} and \ref{subsec:EFit}, respectively, in a more efficient manner. Furthermore, we investigate the advantages of combining these techniques through our proposed ELearnFit approach, which will be described in Subsection \ref{subsec:ELearnFit}.  

To run each model, the input consists of the testing article, which may or may not be accompanied by support article-summary pair samples in the prompt. The output is the generated summary. To generate the summary, we sample from a pre-trained language model using greedy decoding, producing tokens one by one until a stop token is encountered or the maximum token limit of 100 is reached.  To evaluate the performance of the model, the ROUGE-1 F1 score is employed. This metric measures the overlap between the generated summary and the reference summary

Although the main emphasis of this paper is on fine-tuning LLaMa2 models, it is worth highlighting that the strategies and techniques discussed in the following sections can be adapted to optimize the performance of other transformer-based models.

\subsection{Analysis of Data and Performance of Existing Models on Leaderboard}
\label{subsec:Analysis}
\begin{figure}[ht!]
\centering
\includegraphics[width=0.45\textwidth]{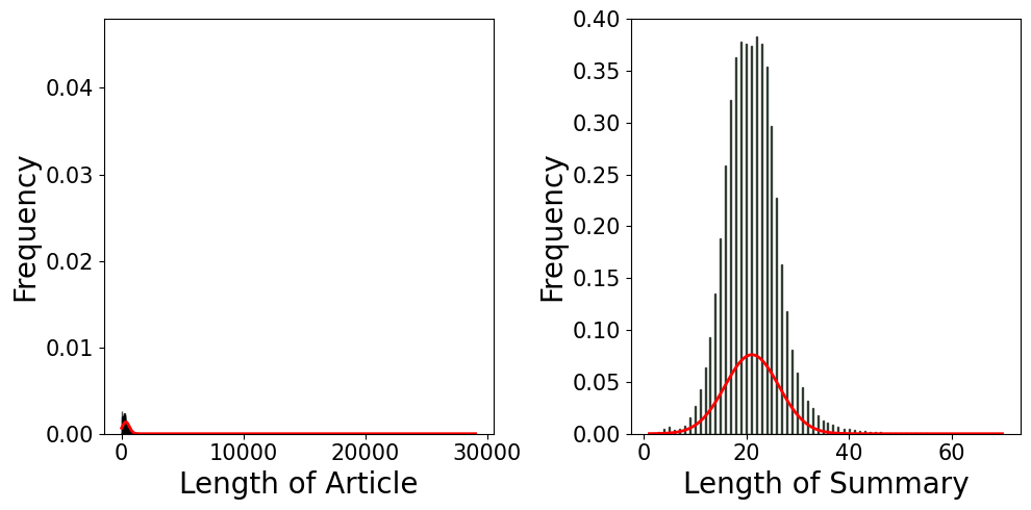}
\caption{Length Distribution of Articles and Summaries in Training Set}
\label{fig:histo_org}
\end{figure}

We conduct our research using the XSum dataset, which consists of a training set comprising 204,045 article-summary samples meticulously curated by the original researchers. In Figure~\ref{fig:histo_org}, the distributions of article and summary lengths in the training set are displayed. 

\begin{figure}[ht!]
\centering
\includegraphics[width=0.45\textwidth]{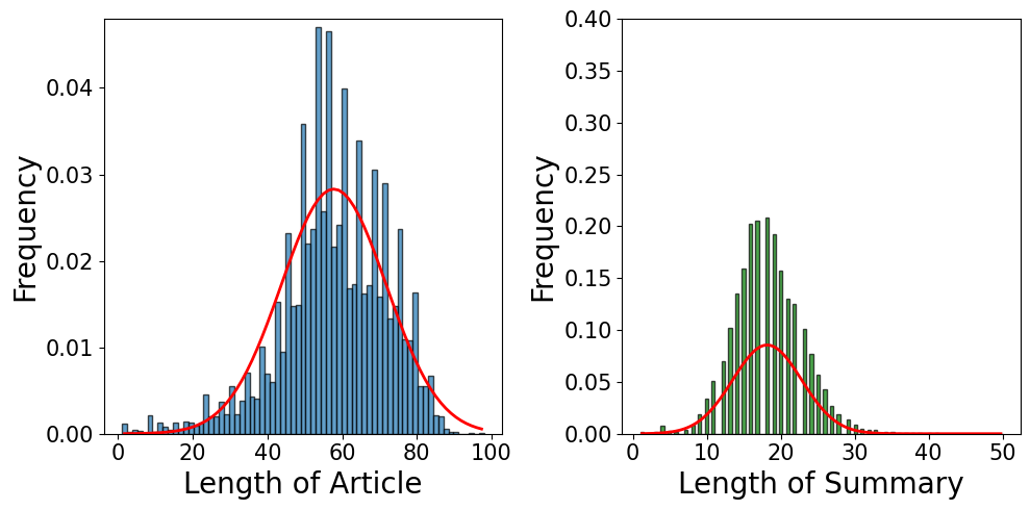}
\caption{Length Distribution of Filtered Articles and Summaries (Combined Word Count $\leq$ 100)}
\label{fig:histo_filtered}
\end{figure}

Due to limited resources, we face constraints (refer to Section \ref{sec:limit} for more details) in using powerful GPU machines to fine-tune models using the complete training dataset. Additionally, the size and input token limits for several representative open-source GPT models, as indicated in Table~\ref{input_limit}, could pose restrictions on testing few-shot learning scenarios.  Consequently, we create a smaller dataset consisting of 17,806 samples from the training set. This subset is obtained by filtering out rows from the training dataset where the combined word count of the article and summary exceeded 100. The length distributions of the filtered articles and summaries are displayed in Figure~\ref{fig:histo_filtered}.  

Based on numerical testing observations, it has been determined that even the filtered dataset is still too large to adequately explore optimal parameters in experiments. To ensure a fair comparison across all experiments, we further reduce the dataset by selecting the initial 256 article-summary pairs as the fine-tuning set. The remaining 125 pairs are reserved for testing purposes.  It's important to note that while the testing set consists of only 125 pairs, the entire filtered dataset (excluding the testing pairs) is utilized to assist the model in selecting relevant support samples for prompting and fine-tuning, as explained in subsections 4.2 and 4.4, respectively. 

According to the leaderboard ranking \cite{XSumLeaderboard}, the top-performing papers in news summarization achieve impressive results by assigning probability mass to candidate summaries \cite{zhao2022calibrating} or by aligning model-generated sequences with reference sequences \cite{liu2022brio}. These approaches consistently yield Rouge-1 scores close to 0.5 across the entire testing dataset. However, in our work, we simply sample from a pre-trained LLM using greedy decoding, generating tokens iteratively until either a stop token is encountered or the maximum token limit of 100 is reached. It is worth noting that our primary focus is on optimizing efficient techniques for in-context learning and fine-tuning in news summarization, with the specific choice of dataset and token adjustment not being crucial to  the outcome of our work.


\begin{table}
\caption{\label{input_limit} Number of Parameters and Input Token Limits for GPT Models (Approximately 1.5 Tokens per Word)}
\centering
\begin{tabular}{|c|c|c|}
\hline
\textbf{Models} & \textbf{Parameters} & \textbf{Input Tokens} \\ \hline
GPT2-Medium &345 million & 1,024 \\
Eleuther-Neo &2.7 billion & 2,048 \\
LLaMa2-7B &7 billion& 2,048\\
LLaMa2-13B &13 billion & 4,096 \\
\hline
\end{tabular}
\end{table}
\subsection{ELearn - Efficient In-Context Learning}
\label{subsec:ELearn}
\begin{figure}[ht!]
\centering
\includegraphics[width=0.48\textwidth]{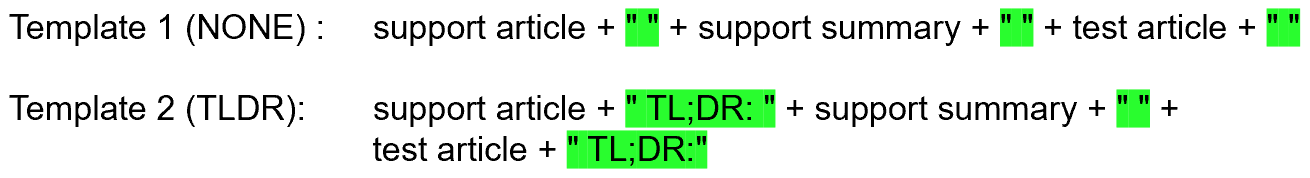}
\caption{Templates for One-Shot Learning: "none" vs "TL;DR"}
\label{fig:template}
\end{figure}

We use two simple templates to investigate the impact of templates on few-shot learning. Figure~\ref{fig:template} illustrates the case for one-shot learning. The first template, called "NONE," utilizes a single space to separate the support article, support summary, and the test article. The second template, known as "TL;DR" (Too Long;Didn't Read), utilizes " TL;DR: " to differentiate between the article and summary (Please note that there intentionally exists a space before and after "TL;DR:" in most occurrences, while in the last occurrence of "TL;DR:", there is only one space before "TL;DR:" and no space after the colon. This formatting choice has been made to facilitate word generation using a language model.). Additionally, a single space is used to separate the support sample from the test sample.  For clarity, these separators are highlighted in green in the figure. 

Figure~\ref{fig:template} presents an example of a one-shot learning template. An interesting aspect to explore is the impact of different numbers of support examples in the prompt. When selecting examples, one approach is to randomly choose article-summary pairs from the training set, which generally provides diversified support examples. Another approach is to use retrieve similar pairs to a given testing article in the prompt, which may result in examples concentrated around specific content or topics.  Furthermore, it is important to consider the size of language models, as it directly relates to memory usage and can potentially influence in-context learning.

\subsection{EFit - Efficient Fine-Tuning}
\label{subsec:EFit}
In LLaMa2 \cite{touvron2023llama}, the transformer block plays a crucial role in the transformer architecture. It comprises of two main sub-layers: a self-attention layer and a feed-forward network. To construct a Transformer model, multiple Transformer blocks are repeated (32 for LLaMa2-7b and 40 for LLaMa2-13b) and stacked together. Each block processes the output of the previous block, allowing the LLaMa2 model to capture both local and global dependencies in the input sequence. However, due to the large size of the model and limited GPU resources, one approach for parameter-efficient fine-tuning is to selectively choose a specific transformer block layer, such as the first layer, to fine-tune the pre-trained weight matrix $W_{\ell}^0 \in \mathbb{R}^{d_1 \times d_2}$ to a new arbitrary weight matrix $W_{\ell}^{ft}$ while freezing the remaining block layers.  

Another approach for parameter-efficient fine-tuning is to employ LoRA (Low-Rank Adaptation). This technique freezes the pre-trained model weights and introduces trainable rank decomposition matrices into each layer of the Transformer architecture. By doing so, the number of trainable parameters for downstream tasks is significantly reduced \cite{hu2021lora}.  Mathematically, LoRA imposes constraints on the fine-tuned parameter space:  $W_{\ell}^{ft} = W_{\ell}^0 + AB^\top$, where $A \in \mathbb{R}^{d_1 \times p}$ and $B \in \mathbb{R}^{d_2 \times p}$ are low rank matrices and $p << d_1, d_2$. With LoRA, the number of parameters being fine-tuned for a single layer is $(d_1 + d_2) \times p$. The original number of parameters for the single layer is $d_1 \times d_2$. Therefore, the ratio of parameters fine-tuned by LoRA to the original parameters is:

$$\frac{(d_1 + d_2) }{(d_1 \times d_2)} \times p = (\frac{1}{d_1} + \frac{1}{d_2}) \times p$$

Let's take the query projection matrix (q\_proj) of the self-attention layer of LLaMa2-7b as an example. The matrix has dimensions of 4,096 x 4,096, with $d_1=4,096$ and $d_2=4,096$. By applying LoRA with a rank parameter of $p=16$ (where $p << d_1$ and $p << d_2$), we achieve a reduction ratio of 0.0078, indicating significant parameter reduction. LoRA proves to be most effective in saving parameters when $p$ is much smaller than both $d_1$ and $d_2$.

Furthermore, inspired by the idea of retrieval augmented generation (RAG) \cite{lewis2021retrievalaugmented}, we incorporate the selection of relevant support examples during the prompting and fine-tuning stages.  This is accomplished by performing semantic search to retrieve top pairs that are similar to each individual testing article. By adopting this approach, the fine-tuning examples can be more targeted and aligned with the specific content or topics covered in the testing articles.  Another alternative is to randomly select article-summary pairs, which introduces a broader range of examples for the fine-tuning process. This random selection provides diverse instances, enhancing the fine-tuned model's robustness and adaptability.

\subsection{ELearnFit - Combine ELearn and EFit}
\label{subsec:ELearnFit}
Both the ELearn and EFit approaches, discussed earlier, have the potential to independently improve model performance. ELearn is preferable when there are few annotated examples available, whereas EFit may be more suitable when numerous examples are accessible. In practice, annotations are costly and often limited to a small number of examples per task. Moreover, training models with a large amount of data necessitates substantial GPU resources and time.  

To address these issues, we propose an approach called ELearnFit, which combines ELearn and EFit by first fine-tuning and then prompting the model. Since both ELearn and EFit have multiple parameters to optimize independently, we employ a heuristic  approach. This involves selecting optimal parameters from the ELearn optimization process and then incorporating the optimal parameters from the EFit optimization process. By doing this, we effectively manage computational resources and time constraints while striving for the best parameter settings. 

For these experiments, prompting is conducted via random sampling of support examples in the prompt to be fed to the pre-trained model. On the other hand, fine-tuning is performed over ten iterations, with data randomly sampled from the training set without replacement in each iteration.  Both the randomly sampled examples in the prompt for ELearn and the fine-tuning process for EFit introduce variability in the fine-tuning process. To comprehensively evaluate and analyze the robustness and performance of ELearn, EFit, and ELearnFit, we investigate which component contributes more to the variation. This investigation is crucial for understanding the stability and reliability of each approach across different trials and conditions. Ultimately, it will help us identify the most robust strategy that consistently delivers strong performance in the presence of variability.

\section{Experiments}
\label{sec:Experiments}
All experiments are run on the Azure ML platform,  harnessing the computational capabilities of A100 GPUs equipped with 80 gigabytes of high-bandwidth memory. This technological foundation provide the ideal setting for a series of groundbreaking investigations.  In order to ensure a systematic exploration and refinement of the parameters, we conduct all experiments in a sequential manner. Through employing a heuristic sequential approach, we efficiently manage computational resources and time constraints while striving for optimal parameter settings.

Subsection \ref{subsec:ELearntest} focuses on ELearn, and compares the results of varying LLM model size, prompt templates, and few-shot learning paradigms.  Subsection \ref{subsec:ELearn_rag} delves into the impact of selective samples for prompting on ELearn.  Subsection \ref{subsec:EFittest} shifts to EFit, and explores the effectiveness of parameter-efficient fine-tuning through two distinct approaches: selective layer finetuning and LoRA algorithms. Subsection \ref{subsec:EFit_ss} sheds light on the insights gleaned from selective training samples for EFit.  Subsection \ref{subsec:ELearnFittest} analyzes the impact of combining the capabilities of ELearn and EFit, resulting in ELearnFit, and highlighting the potential for synergy between these techniques. Lastly, Subsection \ref{subsec:Robustness} compares the robustness of the various models. 

\subsection{Investigate ELearn by Analyzing the Influence of Model Size, Templates, and Few-shot Learning}
\label{subsec:ELearntest}
\begin{figure}[ht!]
\centering
\includegraphics[width=0.45\textwidth]{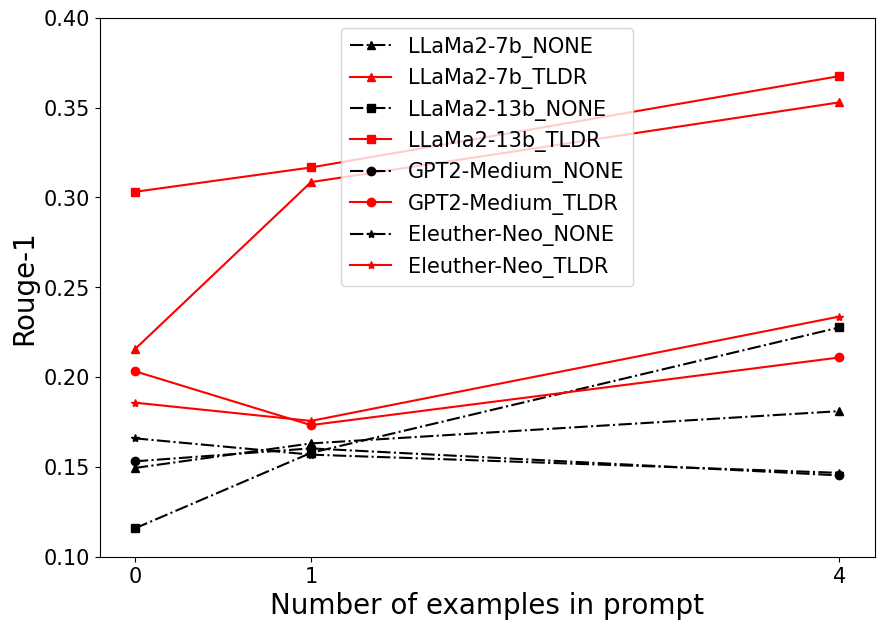}
\caption{Comparison of Four Language Models with Few-shot Learning using Two Templates}
\label{fig:icl}
\end{figure}

In this experiment, we compare four representative open-source GPT models: Eleuther-Neo, GPT2-medium, LLaMa2-7b, and LLaMa2-13b, explore the influence of two prompt templates (none and TL;DR), and vary the number of examples in the prompt. The results are illustrated in Figure~\ref{fig:icl}, where the x-axis represents the number of examples in the prompt and the y-axis represents the Rouge-1 score.

Our findings suggest that increasing the number of examples in the prompt leads to improved model performance. Notably, in the case of GPT-2 models, the zero-shot performance exceeds the one-shot performance. These findings align with previous studies conducted by \cite{brown2020language,xie2022explanation} on datasets such as LAMBADA, HellaSwag, PhysicalQA, and RACE-m, which reported similar observations in relation to GPT-3. 
Additionally, we observe that utilizing a straightforward prompt structure, specifically "TL;DR" (depicted in red), facilitates the model's learning process. This simplified format enables faster pattern recognition in comparison to the none template (depicted in black).  Furthermore, focusing on the four models and examining their performance with the "TL;DR" template (depicted in red) in Figure~\ref{fig:icl}, it becomes evident that LLaMa2-7b and LLaMa2-13b outperform gp2-medium and Eleuther-Neo. This finding suggests that the larger models, LLaMa2-7b and LLaMa2-13b, possess superior capabilities in handling the summarization task, signifying their suitability for this specific application.

\subsection{Enhance ELearn via Selective Samples during Prompting}
\label{subsec:ELearn_rag}
To further improve the perforamcne of ELearn, inspired by the idea of RAG for prompting, we utilize semantic search to retrieve support article-summary samples that are contextually relevant to each testing article, which enables ELearn to learn from these samples in prompts and potentially generate more accurate responses. 

In this experiment, we broaden the range of support samples used in the prompt by including the entire filtered dataset, excluding the samples designated for testing. The outcomes obtained using this expanded scope align closely with those achieved using the original support samples from training set, so we solely showcase the results obtained from the latter (the complete filtered dataset, excluding the 125 testing samples) in this paper.  Note that the order of prompting may potentially lead to different performance results as compared to random prompt ordering. Research conducted by \cite{liu2023lost} demonstrates that in the QA problem, the location of relevant information within the language model's input context follows a U-shaped performance curve. Moreover, the 7B Llama-2 models are biased towards recent information, performing best when it is located at the end of the input context. However, exploring the impact of prompt order for news summerization is beyond the scope of this research paper.

\begin{figure}[ht!]
\centering
\includegraphics[width=0.45\textwidth]{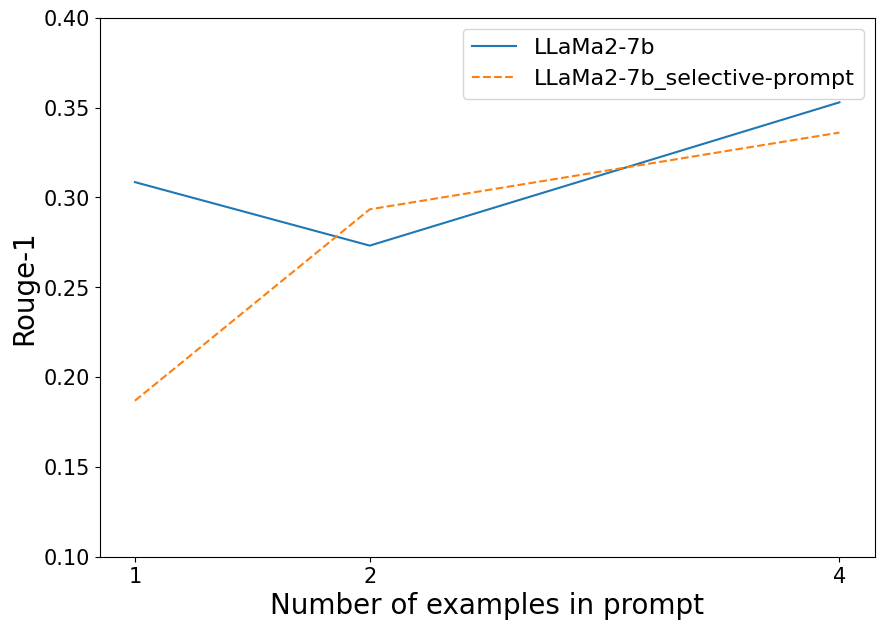}
\caption{An Evaluation of In-Context Learning Methods: Comparing Random Samples vs. Selective Samples in Prompts}
\label{fig:icl_rag}
\end{figure}

Figure~\ref{fig:icl_rag} depicts that the utilization of selective samples during few-shot learning does not significantly affect the performance of the model. One potential explanation for this outcome could be that our straightforward implementation is incapable of capturing the extensive range of topics encompassed in news articles. As a result, the support samples may not adequately represent the diverse range of subjects covered by the articles in the test dataset.

\subsection{Investigate EFit}
\label{subsec:EFittest}
\begin{figure}[ht!]
\centering
\includegraphics[width=0.45\textwidth]{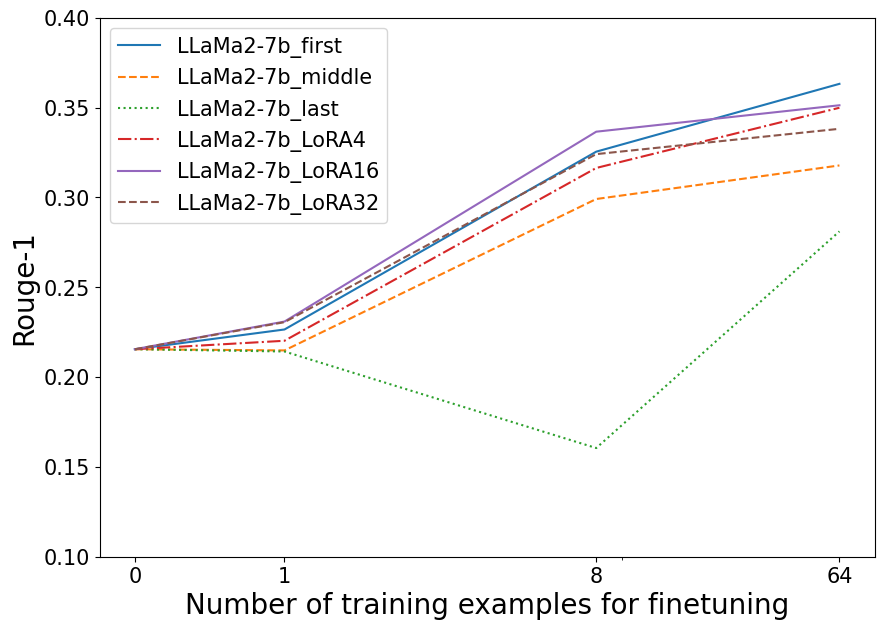}
\caption{Parameter-Efficient Fine-tuning using LoRA and Selective Layer Approaches (Please note that the x-axis is logarithmically scaled for values of the number of support examples greater than 4).}
\label{fig:ft}
\end{figure}

We explore the effectiveness of parameter-efficient fine-tuning using two approaches: LoRA (LoRA4, LoRA16, and LoRA32 algorithms) and selective layers. Figure~\ref{fig:ft} shows the results of the various fine-tuned models with LoRA as well as the models fine-tuned on specific layers (while freezing the remaining layers).

The results suggest that increasing the number of training examples for fine-tuning generally leads to improved performance. When there is only one support example, all algorithms perform similarly. However, with a larger number of support examples (e.g., 8 and 64), fine-tuning the first layer and fine-tuning with LoRA16 results in significantly better performance. Furthermore, when the number of support examples is limited (e.g., 8), fine-tuning the first layer of a LLaMa2-7b often yields weaker results compared to fine-tuning with LoRA16. This is because LoRA16 makes slight modifications to each layer of the LLaMa2-7b, allowing it to adapt more effectively to a small number of examples. However, as the number of support examples increases (e.g., 64), fine-tuning the first layer of the LLaMa2-7b shows improved performance compared to fine-tuning with LoRA16. This is because fine-tuning the first layer allows the LLaMa2-7b to learn task-specific patterns and relationships more directly, leveraging the increased amount of training data.

Additionally, fine-tuning with LoRA16 outperforms both LoRA4 and LoRA32. This suggests that the decomposed weight matrix with a rank of 16 is better suited for representing features learned from news articles compared to ranks 4 and 32.  Finally, the model where only the last layer is fine-tuned performs the worst, suggesting that the pre-trained and fine-tuned data sets do not fully overlap. As a result, fine-tuning the lower-level, granular features proves more effective in improving performance as compared to focusing on high-level features, given an adequate number of support examples.  These findings suggest that fine-tuning the first layer of LLMs has the most impact.

\begin{figure}[ht!]
\centering
\includegraphics[width=0.45\textwidth]{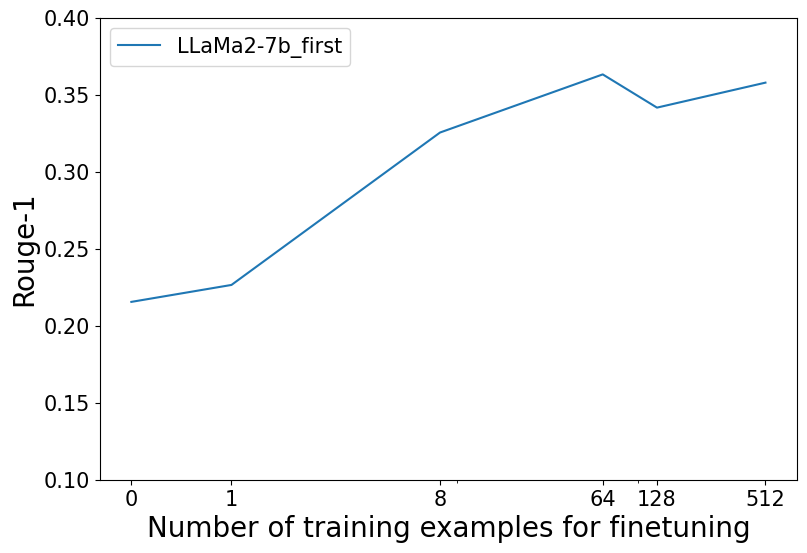}
\caption{Impact of Number of Training Examples on Fine-tuning the First Layer of LLaMa2-7b 
(note that the x-axis has been logarithmically scaled).}
\label{fig:ft_first}
\end{figure}

In practice, annotated examples may not be readily available so we investigate the impact of sample size on model performance. In Figure~\ref{fig:ft_first}, we observe that the Rouge-1 score reaches a local maximum around 64 training examples. Beyond that point, the performance exhibits fluctuations as the number of examples continues to increase. This finding suggests that 64 training examples could potentially represent a "sweet spot" for fine-tuning.
\subsection{Enhance EFit via Selective Training Samples during Fine-tuning}
\label{subsec:EFit_ss}
To enhance the performance of EFit, we draw inspiration from the concepts of selecting relevant samples in the prompting phase. In this experiment, for each testing sample, we select the top 1 or top 2 most similar training samples from the entire filtered dataset, excluding the 125 testing samples. We then fine-tune the model using these selected samples.

Table~\ref{elearn_ss} shows the results.  When fine-tuning the first layer of LLaMa2-7b, using the more similar samples during fine-tuning did not impact model performance. On the other hand, when the model is fine-tuned LoRA16, using the more similar samples led to slightly improved performance. Interestingly, the improved results under LoRA16 are comparable to the results under the model with the fine-tuned first layer. This suggests that the LoRA16 model may benefit from having more relevant samples during fine tuning.
\begin{table}
\caption{\label{elearn_ss} Comparison of Rouge-1 (\%) between EFit with Random Samples and Selective Samples}
\renewcommand{\arraystretch}{1.2}
\centering
\small
\begin{tabular}{|p{3cm}|p{2.2cm}|p{2.2cm}|}
\hline
 \textbf{EFit Sampling} & \textbf{LLaMa2-7b, Fine-tuned First layer} & \textbf{LLaMa2-7b, LoRA16} \\ \hline
Random Sample & 36.32 & 32.43\\
Top 1 Selective Sample & 35.36 & 36.16\\
Top 2 Selective Samples & 36.62 & 34.38\\
\hline
\end{tabular}
\end{table}

\subsection{ELearnFit - Optimize LLM by Combining ELearn and EFit}
\label{subsec:ELearnFittest}
We now look to combine the ELearn and EFit approaches to gain the benefits of both better prompting and fine-tuning. In this experiment, we focus on the TL;DR template in the prompt and two fine-tuned models (fine-tune the first layer of LLaMa2-7b or LLaMa2-7b with LoRA16).

During each testing phase, we first fine-tune LLaMa2-7b and then apply few-shot in-context learning using different numbers (referred to as shots) of support examples  (e.g., 0, 1, 2, 4, and 8 shots). The examples for in-context learning were randomly selected from the training set and incorporated into the prompts.  

\begin{figure}[ht!]
\centering
\includegraphics[width=0.45\textwidth]{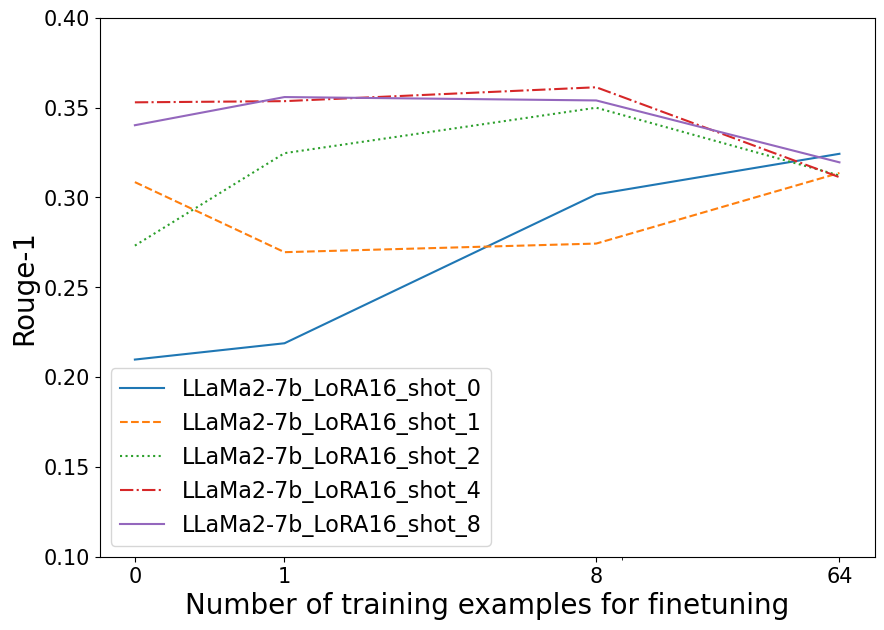}
\caption{Fine-tuning LLaMa2-7b with LoRA16 and Applying Few-shot In-context Learning}
\label{fig:icl_ft_lora}
\end{figure}

\begin{figure}[ht!]
\centering
\includegraphics[width=0.45\textwidth]{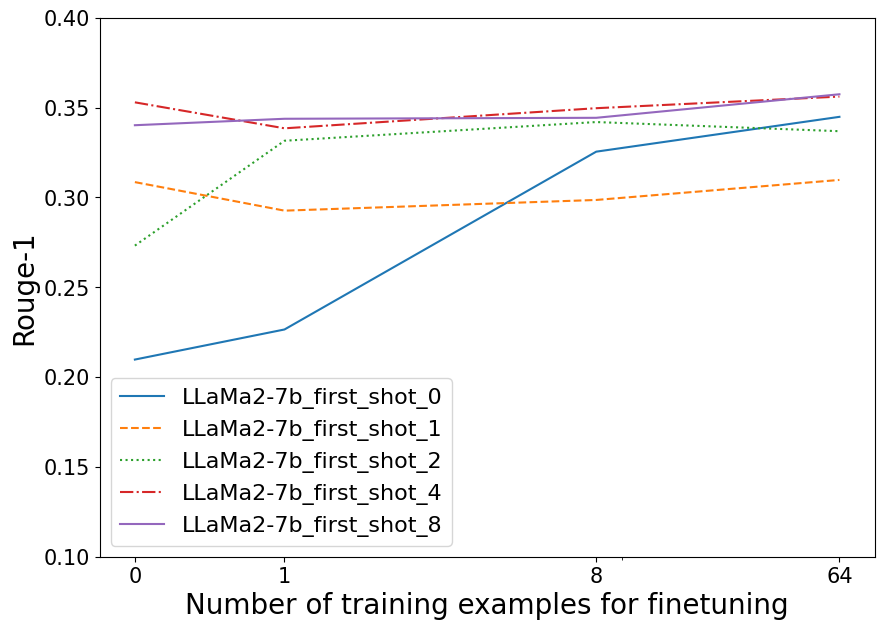}
\caption{Fine-tuning the First Layer of LLaMa2-7b and Applying Few-shot In-context Learning}
\label{fig:icl_ft_first}
\end{figure}

The results, as depicted in Figure~\ref{fig:icl_ft_lora} and Figure~\ref{fig:icl_ft_first}, indicate that when there are limited annotations available for fine-tuning LLaMa2-7b, 4-shot learning leads to superior performance when compared to the results using less shots. Interestingly, both 4-shot and 8-shot learning exhibit similar performance levels. However, this performance gap disappears when there are enough examples for fine-tuning and the results with different shot learnings converge. This suggests that few-shot learning has a lesser impact when a model is effectively fine-tuned with an adequate number of examples. Said another way, having more examples in the prompt can compensate for smaller sample sizes during the fine-tuning process.

\begin{figure}[ht!]
\centering
\includegraphics[width=0.45\textwidth]{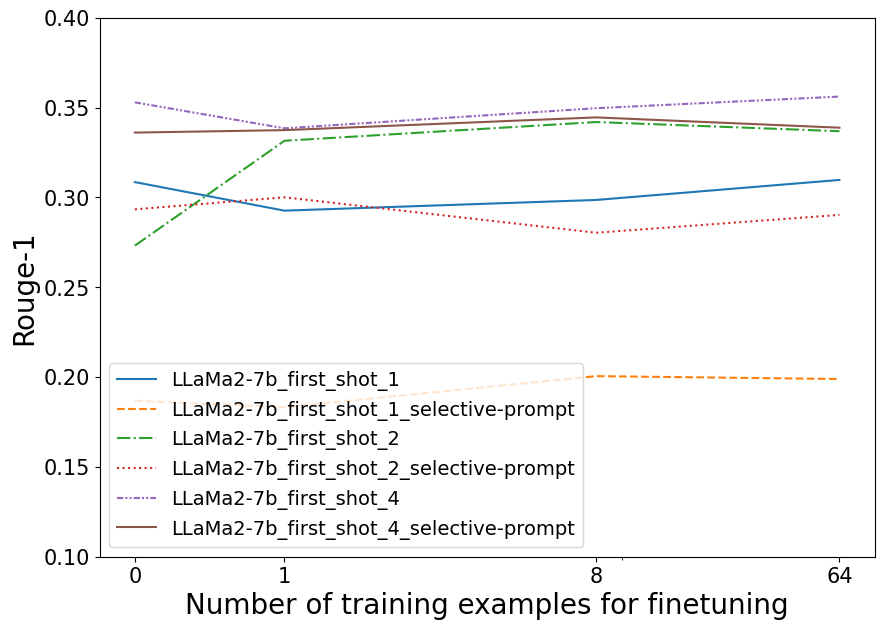}
\caption{Comparing In-Context Learning Approaches: Random Sampling vs. Selective Sampling during Prompting, Following Fine-tuning the First Layer of LLaMa2-7b}
\label{fig:rag_first}
\end{figure}

\begin{figure}[ht!]
\centering
\includegraphics[width=0.45\textwidth]{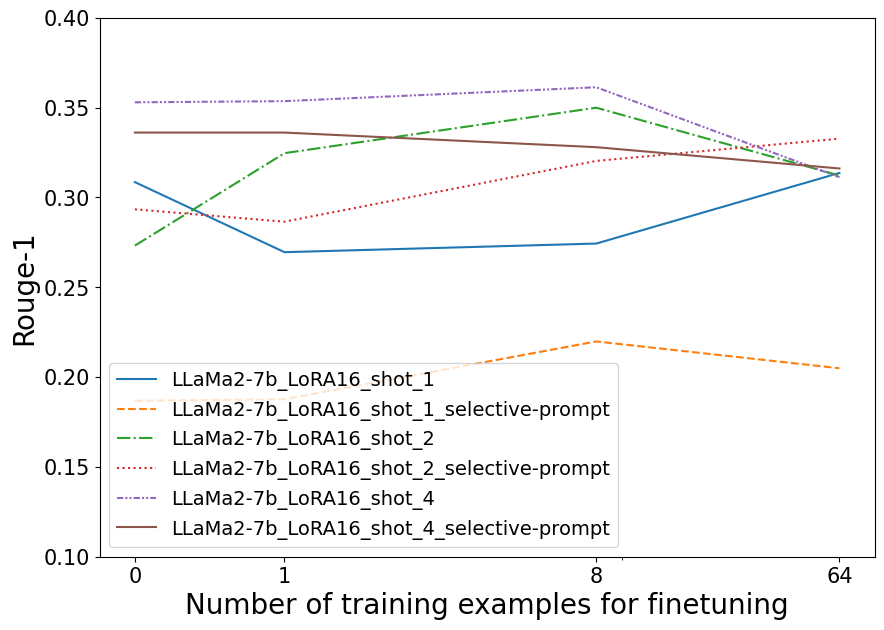}
\caption{Comparing In-Context Learning Approaches: Random Sampling vs. Selective Sampling during Prompting, Following Fine-tuning LLaMa2-7b with LoRA16}
\label{fig:rag_lora}
\end{figure}

\begin{figure}[ht!]
\centering
\includegraphics[width=0.45\textwidth]{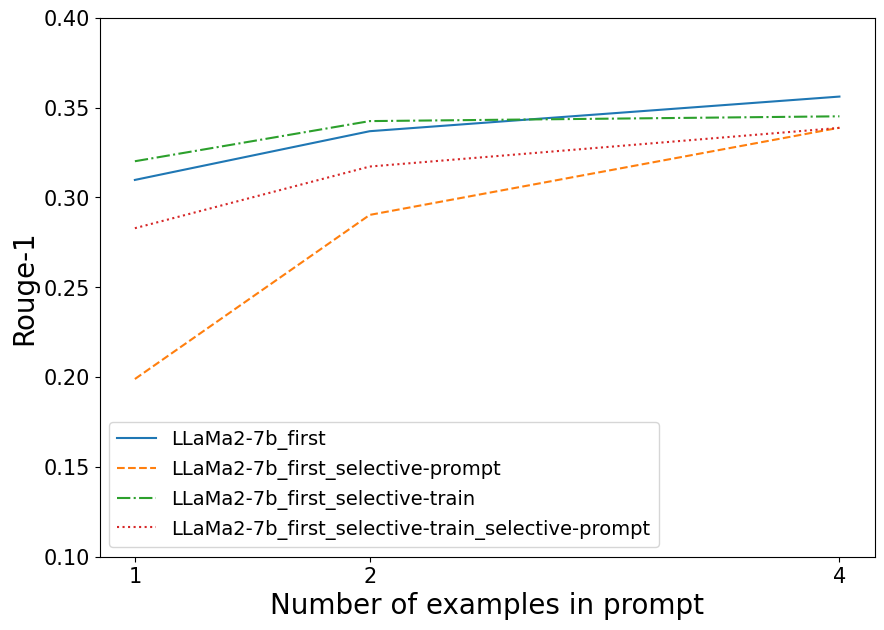}
\caption{Comparing Fine-tuning the First Layer of LLaMa2-7b: Random Sampling vs. Selective Sampling in Training Set, and Random Sampling vs. Selective Sampling during Prompting}
\label{fig:ss_first}
\end{figure}

\begin{figure}[ht!]
\centering
\includegraphics[width=0.45\textwidth]{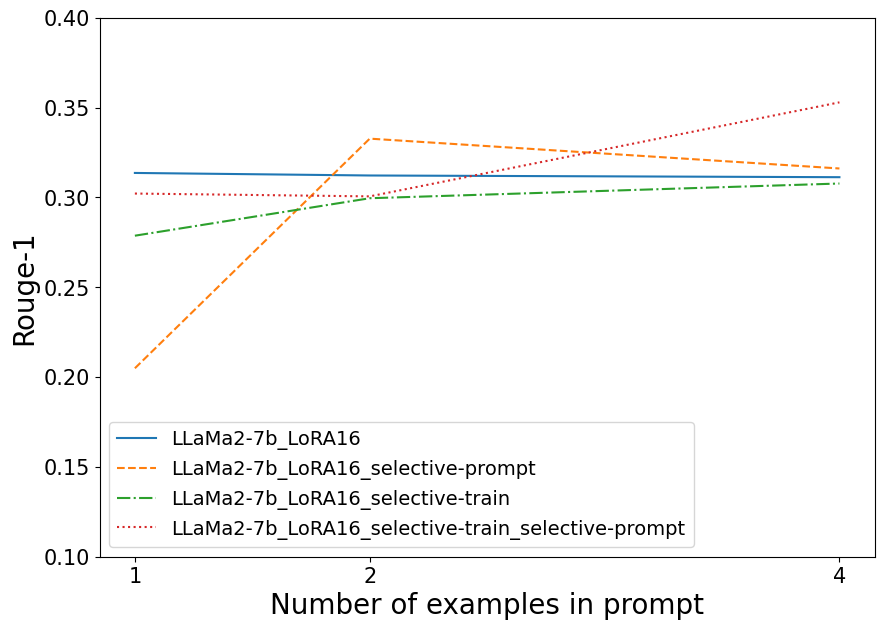}
\caption{Comparing Fine-tuning LLaMa2-7b with LoRA16: Random Sampling vs. Selective Sampling in Training Set, and Random Sampling vs. Selective Sampling during Prompting}
\label{fig:ss_lora}
\end{figure}

Similar to our investigation of selecting relevant samples in few-shot learning for ELearn, we now test whether this approach would benefit ELearnFit during its few-shot learning phase.  Figures~\ref{fig:rag_first} and ~\ref{fig:rag_lora} show the results of applying for ELearnFit after fine-tuning the first layer of LLaMa2-7b and fine-tuning with LoRA16, respectively. It is worth mentioning that when the number of training examples for fine-tuning is zero, it signifies pure in-context learning.  Consistent with our findings in Section 3.2, these results suggest that randomly sampled examples offer a wider range of styles for the LLMs to effectively learn the summarization task. On the other hand, selective sampling faces challenges in capturing the desired diversity.  Furthermore, it is worthwhile noting that when the model undergoes fine-tuning with LoRA16 and has an adequate number of examples (e.g., 64 examples), selective sampling demonstrates a slight improvement in overall model performance.
We now use semantic search to identify the most similar training samples to fine-tune the model.  Figure~\ref{fig:ss_first} shows that fine-tuning the first layer, using selective samples in training and 4-shot learning during in-context learning exhibits slightly inferior performance as compared to the proposed combined approach, which involves 64 examples for fine tuning and 4 examples in prompting. However, it outperforms the ELearnFit approach with 1- or 2-shot learning.  Additionally, as depicted in  Figure~\ref{fig:ss_lora}, when fine-tuning with LoRA16, the combination of selective samples in training, and few-shot learning with four selective samples during in-context learning yields the overall best result. One possible explanation is that the use of selective samples for finetuning for prompting together could potentially enhance the effectiveness of finetuning LLaMA2-7b with LoRA16. This proposition finds support in the comparison between Figure~\ref{fig:icl_ft_lora} and Figure~\ref{fig:icl_ft_first}. Specifically, when evaluating the performance of the 4-shot learning scenarios, an increase in the number of examples for finetuning from 8 to 64 results in a degradation in performance for the former, as depicted in Figure~\ref{fig:icl_ft_lora}. In contrast, the latter exhibits a stable performance, as illustrated in Figure~\ref{fig:icl_ft_first}.  

\subsection{Robustness Checks}
\label{subsec:Robustness}
\begin{figure}[ht!]
\centering
\includegraphics[width=0.45\textwidth]{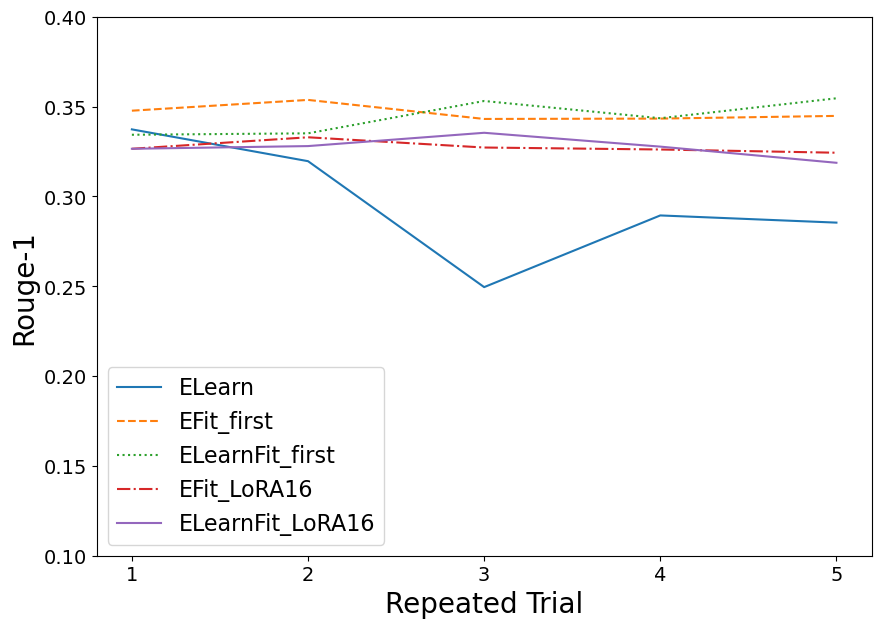}
\caption{Robustness Comparison of ELearn, EFit and ELearnFit}
\label{fig:robustness}
\end{figure}

In our experiment, fine-tuning is performed over ten iterations. In each iteration, data are randomly sampled from the training set without replacement, introducing variability in the fine-tuning process. We now assess the robustness of three approaches: ELearn, EFit, and ELearnFit. The descriptions for each model are detailed in Table~\ref{robust_models}.  Figure~\ref{fig:robustness} presents the results obtained from five repeated trials for each approach. The x-axis represents the nth trial, while the y-axis displays the Rouge-1 score.  While we were limited to five trials due to computational constraints, additional trials could be conducted to further assess the robustness of these approaches.  This experimental setup allowed us to gain insights into the performance of each approach under varying conditions and to compare their effectiveness in different scenarios.

Table~\ref{robust_ms} reveals that in-context learning exhibits greater variability across trials compared to the other two approaches. This is evident from the higher standard deviation observed in the ELearn results. In contrast, both EFit\_first and ELearnFit\_first demonstrated similar performance, although ELearnFit\_first had twice the standard deviation of EFit\_first. A similar observation can be made for ELearnFit\_LoRA16 and EFit\_LoRA16.  These findings further suggest that fine-tuning offers more stable performance than in-context learning. Additionally, when the number of samples for fine-tuning is limited, the combined approach ELearnFit yields consistent and reliable performance across different trials, highlighting its potential for enhancing robustness.

\begin{table}
\renewcommand{\arraystretch}{1.2}
\centering
\raggedright
\small
\caption{\label{robust_models} Model Description for Robustness Comparison}
\begin{tabular}{|p{2.5cm}|p{1.8cm}|p{3cm}|}
\hline
\textbf{Model} & \textbf{In-context Learning} & \textbf{Fine-tuning} \\ \hline
ELearn & 4 Shots & - \\
EFit\_first & - & First Layer w/ 64 Examples \\
EFit\_LoRA6 & - & LoRA16 w/ 64 Examples \\
ELearnFit\_first & 4 Shots  & First Layer w/ 64 Examples \\
ELearnFit\_LoRA16 & 4 Shots & LoRA16 w/ 64 Examples \\
\hline
\end{tabular}
\vspace{-7pt} 
\end{table}

\begin{table}
\renewcommand{\arraystretch}{1.2}
\raggedright
\small
\caption{\label{robust_ms} Performance Details for Robustness Comparison}
\begin{tabular}{|p{2.5cm}|p{2.4cm}|p{2.4cm}|}
\hline
\textbf{Model} & \textbf{Mean}& \textbf{Standard Deviation} \\ \hline
ELearn & 0.2962 & 0.0303\\
EFit\_first & 0.3465 & 0.0039\\
EFit\_LoRA16 & 0.3274 & 0.0029\\
ELearnFit\_first & 0.3441 & 0.0086\\
ELearnFit\_LoRA16 & 0.3273 & 0.0053\\
\hline
\end{tabular}
\end{table}

\section*{Limitations} 
\label{sec:limit}
In this paper, we primarily directed our attention to the LLaMa2-7b model, a formidable language model consisting of 7 billion parameters. Assuming that each parameter occupies a modest 4 bytes of memory, the estimated total memory requirement for this model is approximately 27.34 gigabytes, calculated as follows:
\begin{multline}
\text{Total Memory Size} = 7 \times 10^9 \times 4 \text{ bytes} / (1024^2) \\
\approx 27.34 \text{ gigabytes}
\end{multline}
\text{where:}
\begin{align*}
1 \text{ kilobyte (KB)} &= 1024 \text{ bytes} \\
1 \text{ megabyte (MB)} &= 1024 \text{ kilobytes}
\end{align*}
Similarly, the total memory requirements for the LLaMa2-13b and LLaMa2-70b models are approximately 51 gigabytes and 274 gigabytes, respectively.  Due to limited resources on A100 GPUs, which offer up to 80 gigabytes of high-bandwidth memory, and the substantial computation time required for each experiment, we primarily focus on optimizing ELearn and EFiT with the LLaMa2-7b model in this paper. However, we believe that the insights gained from this research work can be readily extended to larger language models such as LLaMa2-70b, especially when coupled with more powerful GPU resources.

\section*{Conclusion}
\label{sec:Conclusion}
News summarization has become increasingly important as the volume of information has exploded. In our research, we explore different techniques to enhance news summaries. Under prompting (ELearn), we demonstrate that using larger models, adding more prompts, and utilizing simple templates improve performance.  We also show that fine-tuning (EFit) enhances performance, especially when the first layer of models is fine-tuned. 

Surprisingly, for both prompt engineering and fine-tuning, leveraging more relevant samples does not improve performance. This is likely due to the fact that news articles are very diverse, and retrieving highly relevant samples during prompting or fine-tuning may result in over-learning, resulting in the model's failure to adequately capture the wide range of topics covered in the test dataset.

Finally, we show that our combined model (ELearnFit) produces the best performance, particularly for situations where there are few annotated samples. In practice, our research suggests that a fine-tuned model (especially on the first layer) coupled with diverse examples during prompting, yields optimal performance for news summarization.


\bibliographystyle{ACM-Reference-Format}
\bibliography{reference}


\end{document}